\documentclass{article}

\usepackage[nonatbib,final]{nips_2017}

\usepackage[utf8]{inputenc} % allow utf-8 input
\usepackage[T1]{fontenc}    % use 8-bit T1 fonts
\usepackage{hyperref}       % hyperlinks
\usepackage{url}            % simple URL typesetting
\usepackage{booktabs}       % professional-quality tables
\usepackage{amsfonts}       % blackboard math symbols
\usepackage{nicefrac}       % compact symbols for 1/2, etc.
\usepackage{microtype}      % microtypography
\usepackage{amsmath}
\usepackage{algorithm}
\usepackage[noend]{algpseudocode}

\usepackage[pdftex]{graphicx}

\title{Superhuman Accuracy on the SNEMI3D Connectomics Challenge}

\author{
  Kisuk Lee \\
  MIT \\
  \texttt{kisuklee@mit.edu} \\
  \And
  Jonathan Zung \\
  Princeton University \\
  \texttt{jzung@princeton.edu} \\
  \And
  Peter Li\\
  Google\\
  \texttt{phli@google.com}\\
  \And
  Viren Jain\\
  Google\\
  \texttt{viren@google.com}\\
  \And
  H. Sebastian Seung \\
  Princeton University \\
  \texttt{sseung@princeton.edu} \\
}

\begin{document}

\maketitle

\begin{abstract}

For the past decade, convolutional networks have been used for 3D reconstruction
of neurons from electron microscopic (EM) brain images. Recent years have seen
great improvements in accuracy, as evidenced by submissions to the SNEMI3D
benchmark challenge.  Here we report the first submission to surpass the
estimate of human accuracy provided by the SNEMI3D leaderboard.  A variant of 3D
U-Net is trained on a primary task of predicting affinities between nearest
neighbor voxels, and an auxiliary task of predicting long-range affinities. The
training data is augmented by simulated image defects. The nearest neighbor
affinities are used to create an oversegmentation, and then supervoxels are
greedily agglomerated based on mean affinity.  The resulting SNEMI3D score
exceeds the estimate of human accuracy by a large margin.  While one should be
cautious about extrapolating from the SNEMI3D benchmark to real-world accuracy
of large-scale neural circuit reconstruction, our result inspires optimism
that the goal of full automation may be realizable in the future.

\end{abstract}

%-------------------------------------------------------------------------------
% Section 1. Introduction
%-------------------------------------------------------------------------------
\section{Introduction}
\label{intro}

The 3D reconstruction of neurons from electron microscopic (EM) brain images is
a basic computational task in the field of connectomics \cite{plaza2014toward}.
Ten years ago it was first demonstrated that convolutional networks could
outperform other image segmentation algorithms at the task
\cite{jain2007supervised}. Recently the DeepEM3D convolutional
net~\cite{zeng2017} approached human accuracy for the first time on the SNEMI3D
benchmark challenge\footnote{\url{http://brainiac2.mit.edu/SNEMI3D/home}} for
segmentation of EM brain images.

Here we describe our own SNEMI3D submission, which is currently at the top of
the leaderboard and has surpassed the SNEMI3D estimate of human accuracy by a
large margin.  Our submission is a variant of U-Net \cite{ronneberger2015} and
differs from other leading SNEMI3D entries \cite{zeng2017,beier2017} by making
more extensive use of 3D convolution.  For realizing the full power of 3D, we
have found two tricks to be helpful.  First, we introduce novel forms of
training data augmentation based on simulation of known types of image defects
such as misalignments, missing sections, and out-of-focus sections.  Second, we
train the convolutional net to predict affinities of voxel pairs that are
relatively distant from each other, in addition to affinities of nearest
neighbor voxels.  In quantitative and qualitative comparisons, we find that both
tricks produce substantial improvements in the performance of convolutional
nets.

That being said, convolutional nets are typically just one stage of an image
processing pipeline, and it is also important to assess the overall accuracy of
the pipeline.  For example, test-time augmentation with rotations and
reflections has been shown to enhance segmentation accuracy \cite{zeng2017}.
This ensemble technique presumably averages out noise in boundary predictions,
at the cost of drastic increase in inference time.  Here we instead apply mean
affinity agglomeration as a postprocessing step, and show that it yields a
comparably large gain in segmentation accuracy, while being much faster to
compute.

The gain from either postprocessing technique (test-time augmentation or mean
affinity agglomeration) is larger for our worst nets than for our best nets.  In
other words, the effects of our two training tricks are reduced by
postprocessing, though they are not eliminated. In particular, for our best net,
the improvement from postprocessing is relatively small.  It is possible that
future progress in convolutional nets will render both postprocessing techniques
ineffective.

Although our SNEMI3D submission has surpassed the estimate of human accuracy
provided by the SNEMI3D leaderboard, one should not jump to the conclusion that
the problem of automating neuronal reconstruction has been completely solved. A
human expert can still readily find mistakes in our submission, if provided with
interactive 3D viewing of segments in addition to 2D views of images and
segmentation. This may seem inconsistent with the SNEMI3D estimate of human
accuracy, unless one appreciates that human accuracy is somewhat ill-defined.
Humans vary greatly in their expertise at the task. Furthermore, accuracy
depends on the procedures and software tools used to perform the reconstruction.
(It is much more difficult for a human expert to find errors in our SNEMI3D
submission based on inspection of the EM images alone, without access to 3D
renderings of the segments.) Therefore it would be a mistake to conclude that
the segmentation problem is now solved. The correct conclusion is that the
SNEMI3D challenge has become obsolete in its present form, and must be modified
or replaced by a challenge that is capable of properly evaluating algorithms
that are now exceedingly accurate.

Having mentioned these caveats, it seems safe to say that the prospects for full
automation of neural circuit reconstruction look more encouraging than ever
before.

%% \paragraph{3D ConvNets for connectomic reconstruction} Seung Lab
%% works~\cite{turaga2009,turaga2010,helmstaedter2013,kim2014}.
%% SegEM~\cite{berning2015}. SyConn~\cite{sven2017}. Both are based on skeleton
%% reconstruction combined with 3D convolutional networks.

%% \paragraph{DeepEM3D} The former leading entry on SNEMI3D ~\cite{zeng2017}. An
%% ensemble of three fully convolutional networks based on the Inception
%% architecture~\cite{szegedy2014}, each taking different number of input slices
%% (1, 3, 5). Only few 3D convolutions in its initial layers. Test-time
%% augmentation with 16 variants for each model, thus blending 48 outputs.
%% Specially treating misalignment errors in the test volume.

%% \paragraph{Multicut} Formly second place on SNEMI3D~\cite{beier2017}. Used a 2D
%% fully convolutional network based on the Inception architecture, with test-time
%% augmentation of 20 variants. The network takes as input three adjacent slices
%% but it is uncertain whether 3D convolution was employed. Multicut and lifted
%% multicut agglomeration.

%% \paragraph{Flood-filling network} Completely different approach to connectomic
%% reconstruction by~\cite{januszewski2016}. Recurrent application of 3D
%% convolutional network. Pros: end-to-end, cons: costly inference. Its
%% applicability to highly anisotropic images of cortex acquired from serial
%% section EM has not been proven yet.

%-------------------------------------------------------------------------------
% Section 3. Residual Symmetric U-Net
%-------------------------------------------------------------------------------
\section{Residual Symmetric U-Net}
\label{RSUNet}

\subsection{Network architecture}

%For instance, a U-Net variant equipped with residual skip connections~\cite{he2015b}
%has recently been proposed independently by multiple research groups
%~\cite{milletari2016,fakhry2016,quan2016,yu2017}. These architectures differ
%only in small details but share common design principles, thus leading us to
%believe that having different names for each architecture may cause unnecessary
%confusion.

%Although it is accurate to say that our network is a 3D version of
%FusionNet~\cite{quan2016}, or a fully residual convolutional network in their
%language, we think U-Net deserves the most credit for its pioneering
%architectural innovation. Therefore, we have decided to emphasize the fact that
%U-Net is directly ancestral to our network by explicitly calling it a %``residual
%symmetric U-Net''. Detailed description of our network architecture will follow,
%elucidating why it is \textit{symmetric} and \textit{residual}.

% Architecture.
\begin{figure}[!t]
\begin{center}
\includegraphics[width=1.0\textwidth]{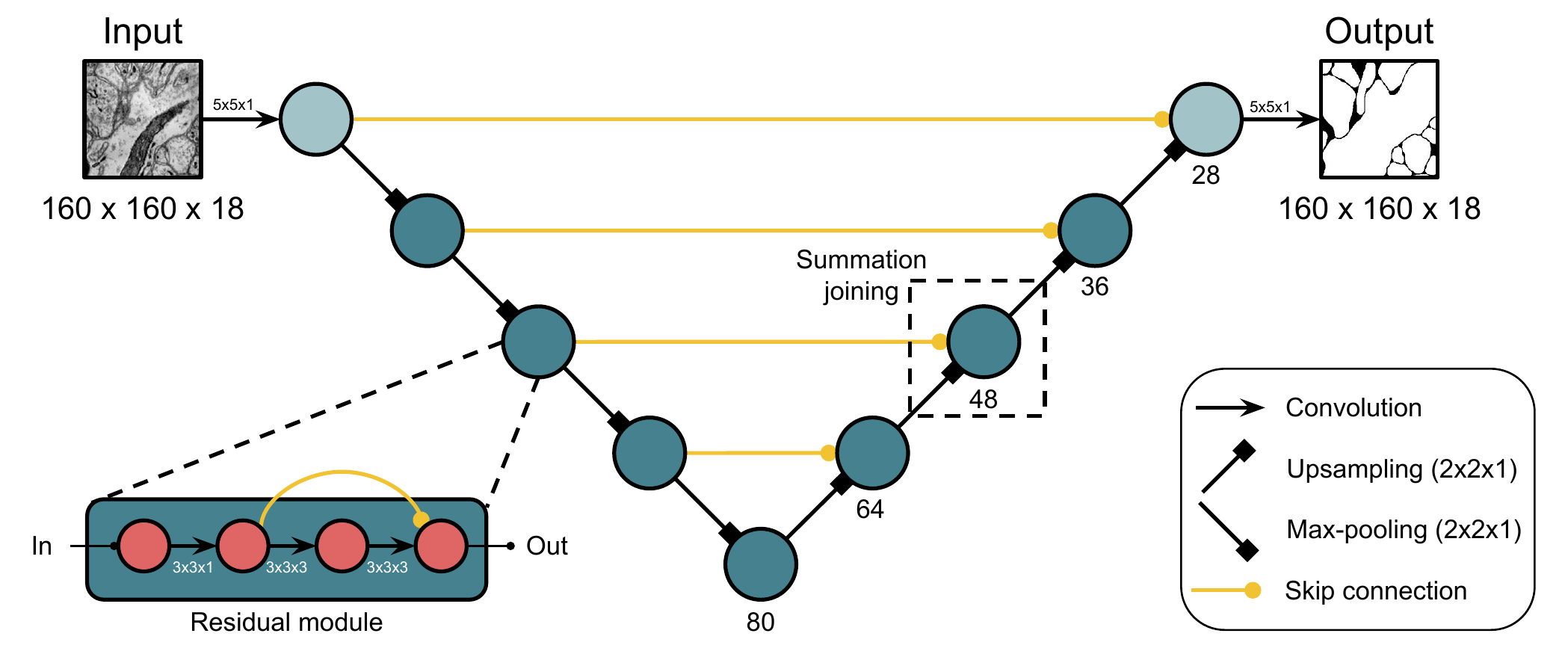}
\end{center}

\caption{Residual Symmetric U-Net architecture. Upsampling is implemented with
strided transposed convolution~\cite{dumoulin2016}, and downsampling with
max-pooling. The numbers below the modules represent the width (or number of
feature maps) at each scale. The light-colored modules at the finest scale (top
level) indicate they exclusively contain 2D convolutions. Detailed description
is presented in Section~\ref{RSUNet}.}

\label{fig:architecture}
\end{figure}

Our network is a variant of the widely used U-Net~\cite{ronneberger2015}. The
architecture (Figure~\ref{fig:architecture}) inherits three main elements from
U-Net: (1) a contracting path with convolutions and downsampling, (2) an
expanding path with convolutions and upsampling, and (3) same-scale skip
connections from the contracting path to the expanding path. These three
elements constitute a top-down refinement process~\cite{pinheiro2016,lin2016} by
progressively integrating higher-level contextual information with lower-level
localization information in a coarse-to-fine manner.

\paragraph{Symmetric architecture} Following others~\cite{quan2016}, we have
modified U-Net to use \textit{same} rather than \textit{valid} convolution. This
is mainly for simplicity and convenience; it is easier to keep track of feature
map sizes.  Border effects may hurt accuracy, but this can be mitigated by the
overlap-blend inference scheme described in Section~\ref{inference}. We further
replace \textit{concatenation} joining by \textit{summation}
joining~\cite{quan2016,smith2016} where the skip connections join the expanding
path.

% These two modifications make the contracting/expanding paths more \textit{symmetric},

\paragraph{Modular architecture} The basic module
(Figure~\ref{fig:architecture}) consists of three convolution layers of equal
width, interspersed with batch normalization layers~\cite{ioffe2015} and
exponential linear units~\cite{clevert2015}. Using the same modules everywhere
simplifies the specification of our network~\cite{shen2017}. The \textit{depth},
or the number of layers along the longest path of the network becomes a function
of how many layers the module contains and how many scales the network spans.
The \textit{scale} is determined by the number of up/downsamplings. The
\textit{width}, or the number of feature maps at each scale can be adjusted to
control network's overall capacity.

\paragraph{Residual architecture}  We have added a residual skip
connection~\cite{he2015b} to each module (Figure \ref{fig:architecture}), thus
making every path from the network's input to its output a \textit{residual}
subnetwork~\cite{veit2016}. Residual variants of U-Net were previously applied
to biomedical image segmentation~\cite{milletari2016,yu2017} and serial section
EM image segmentation \cite{quan2016,fakhry2016}.

\paragraph{Anisotropic 3D} A ``fully'' 3D U-Net~\cite{cicek2016} can be
constructed by expanding the 2D filters for convolution and up/downsampling into
3D. To better deal with the high anisotropy of serial section EM images, we have
made three design choices. First, we never downsample feature maps along the
$z$-dimension so as to minimize the loss of information along the $z$-dimension
with inferior quality. Second, we exclusively use 2D convolutions in the modules
at the finest scale (or highest resolution) where anisotropy is maximal
(light-colored nodes, Figure~\ref{fig:architecture}). Third, the modules in
other scales always start with $3\times3\times1$ convolution, followed by two
$3\times3\times3$ convolutions (Figure~\ref{fig:architecture}). With this
particular choice of filter size, each module represents $7\times7\times5$
nonlinear computation, which is slightly anisotropic. Another motivation for
this particular design choice is to embed 2D features first and then refine them
with residuals from 3D context.

\subsection{Inference}
\label{inference}

\paragraph{Blending} Our use of \textit{same} convolutions allows us to use an
output patch of the same size as our input patch, but accuracy is worse near the
borders of the output patch. At test time, we perform inference in overlapping
patches, and blend them using a bump function which weights the center of an
output patch more strongly than its borders, $f(\vec{r})~=~\exp
\left(\sum_{a=x,y,z} \left[r_a(p_a-r_a)\right]^{-t_a}\right)$, where
$r_x,r_y,r_z$ are the local coordinates within patch, $p_x,p_y,p_z$ are the size
of the patch, and $t_x,t_y,t_z$ control how fast the bump function decays from
center to border in each dimension. We used $t_x, t_y, t_z=1.5$ and $50\%$
overlap in all three dimensions.

\paragraph{Test-time augmentation} Test-time augmentation has been widely
adopted as an effective way of improving the quality of segmentation
~\cite{zeng2017,ronneberger2015,beier2017,quan2016}. The most common set of
transformations includes rotations by $90^\circ$ and horizontal/vertical flips
over the $xy$-plane, resulting in 8 variants. Zeng et al.~\cite{zeng2017} also
added a flip in $z$-dimension, increasing the set size to 16. We also used the
same set of transformations (16 variants) when demonstrating the effect of
test-time augmentation. However, we did not use test-time augmentation when
demonstrating the effect of mean affinity agglomeration, which will be described
in Section~\ref{mean_affinity}.

% \paragraph{Watershed segmentation}

%-------------------------------------------------------------------------------
% Section 4. Long-range affinity
%-------------------------------------------------------------------------------

% Affinty graph.
\begin{figure}[!t]
\begin{center}
\includegraphics[width=1.0\textwidth]{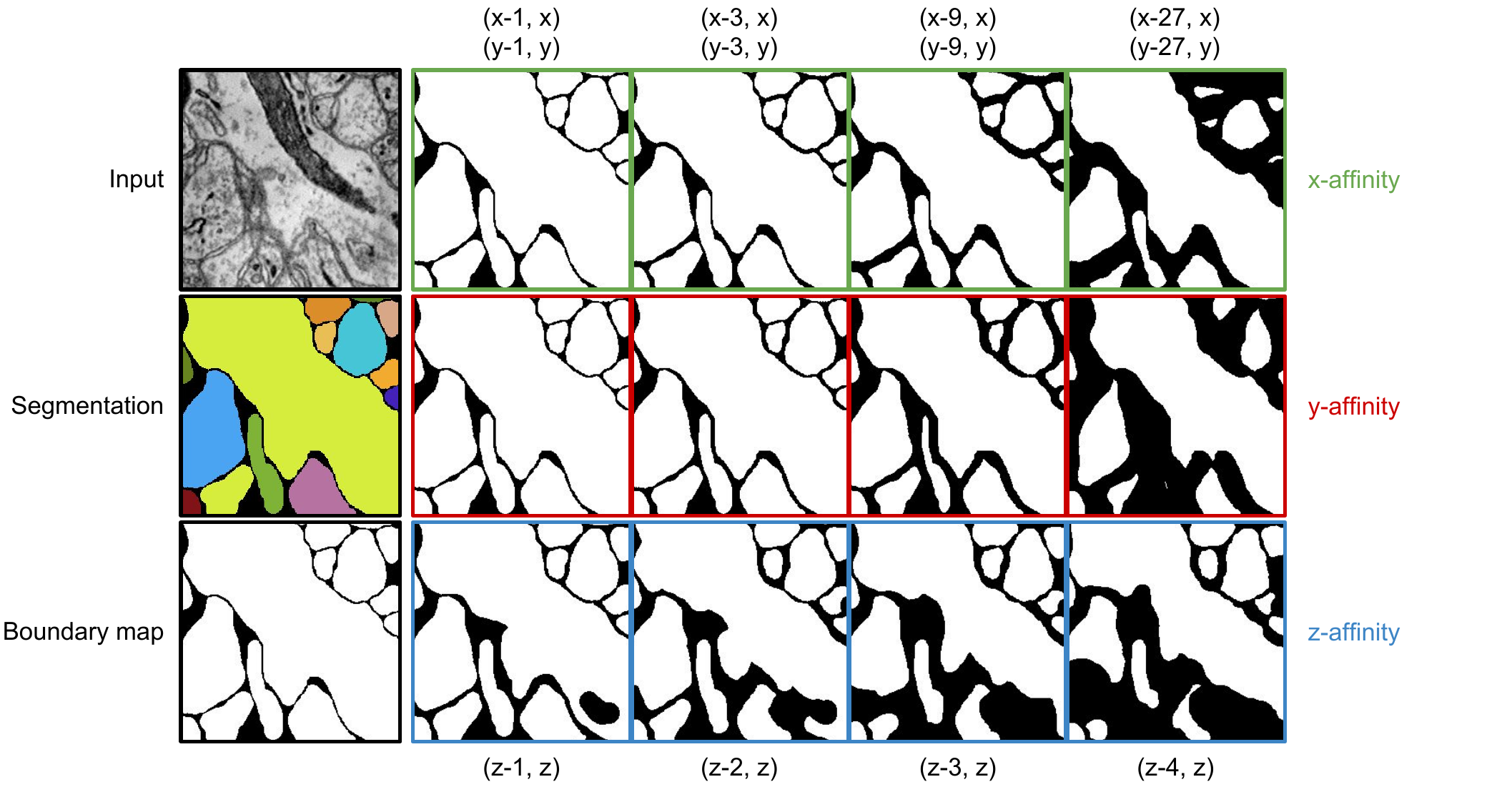}
\end{center}

\caption{An example affinity graph representation. The second column is the
nearest neighbor affinity maps that were used to produce segmentation, and the
third to fifth columns illustrate long-range affinities we introduced as an
auxiliary target to predict. $(a,b)$ represents an undirected edge between the
voxels $a$ and $b$.}

\label{fig:affinity}
\end{figure}

\section{Long-range affinity prediction as an auxiliary task}
\label{long_range}

Turaga et al.~\cite{turaga2010} trained convolutional networks to transform an
input EM image stack into an output affinity graph, which is subsequently
partitioned to produce a segmentation. They included only edges between nearest
neighbor voxels in the affinity graph.  We additionally trained our
convolutional net to predict affinities for a select group of longer edges
oriented along the cardinal directions.  In the $x$- and $y$-directions, the
edges spanned 3, 9, and 27 voxels. In the $z$-direction, the edges spanned 2, 3,
and 4 voxels.

In total, our convolutional net was trained to produce twelve output images, one
corresponding to each of the affinity maps in Figure~\ref{fig:affinity}.  The
long-range affinity maps (third to fifth columns) look qualitatively different
from the nearest neighbor affinity maps (second column).

The \textit{long-range} affinities were not used at test time. They were
included in the training in the hope that they would improve accuracy at the
main task, the prediction of nearest neighbor affinities.  In other words, we
hoped that training on auxiliary tasks would improve performance at the main
task~\cite{zhang2014}.  As Figure~\ref{fig:affinity} shows, the auxiliary tasks
exhibit considerable diversity, which could aid training.

%We hypothesized that our edge-based affinity graph is a better target
%representation than the widely used voxel-based boundary map. First, an affinity
%graph can be viewed as an upsampled version of the boundary map, thus providing
%richer information. Second, $z$-affinity map can provide the networks with
%essential information about the anisotropic nature of the input data. Note that
%$z$-nearest neighbor affinity map is quite different from $x$- and $y$-nearest
%neighbor affinity maps that are similar to the boundary map
%(Figure~\ref{fig:affinity}). Fakhry et al.~\cite{fakhry2016} resorted to the
%thickened version of boundary map to deal with anisotropy, but such a
%manipulation led to losing high-frequency spatial information.

%-------------------------------------------------------------------------------
% Section 5. Data augmentation
%-------------------------------------------------------------------------------
\section{Data augmentation}
\label{augmentation}

Following standard practice, we augmented our training set using random
rotations by $90^\circ$ and flips in $x$-, $y$-, and $z$-dimensions. We also
applied warping and brightness and contrast perturbations using code from
ELEKTRONN \footnote{\url{http://elektronn.org/}}, an open source deep learning
framework. These kinds of augmentation have been widely used when training
convolutional networks on serial section EM
images~\cite{zeng2017,ronneberger2015,beier2017,quan2016}.

We also introduced three novel types of data augmentation. These were motivated
by the necessity of dealing with common image defects: misalignments, missing
sections, and out-of-focus sections. However, we speculate that these kinds of
data augmentation may end up improving accuracy even at locations without image
defects, because they force networks to maximally exploit 3D context.

%to their role in covering the problem space~\cite{smith2016}, we found
%that these augmentations are very effective at guiding 3D networks to
%maximally exploit 3D context. Each data augmentation represents noisy,
%corrupted and/or missing input signals, so 3D context can be very
%useful for recovering from such partial information.

\paragraph{Misalignment} Misalignments of serial section EM images can lead to
severe merge and split errors. Robustness to misalignment is important for
accuracy, though a training set may contain very few examples of misalignment.
To deal with this problem, we introduced a simulated misalignment in every
training sample. Specifically, we picked a random $z$-location in each input
patch and then applied random translations along the $x$- and
$y$-direction~\footnote{The same transformation is applied to both the input
image and target label stacks. The target affinity graph is then dynamically
generated from the transformed label.}. The pixel displacement in each direction
was chosen indepdently from the discrete uniform distribution between 0 and 17.
We generated two different types of misalignment: (1) \textit{slip}-type
misalignment applies the random translation only at the randomly chosen
$z$-location, whereas (2) \textit{translation}-type misalignment additionally
applies the same translation to every slice below the $z$-location.

\paragraph{Missing section} Missing section is another common mode of failure in
serial section EM imaging. In some cases the whole sections are missing, or
sections could be partially missing due to error of imaging. In other cases,
sections become so severely damaged that it is preferable to remove their
content - either partially or fully. When reconstructing large image volumes,
accounting for these errors can be critical for performance. Since our training
set did not contain any missing section, we introduced \textit{missing section}
augmentation. We picked random $z$-locations up to five slices in each input
patch and introduced partial or full missing section independently. We found
that filling out missing sections with zero intensity values distorted the input
distribution too much and damaged inference performance when paired with batch
normalization. Therefore, we drew random fill-out values uniformly from minimum
(zero) to maximum (one) intensity.

\paragraph{Out-of-focus section}  During automated EM imaging, the microscope
focus may occasionally fail and yield blurry images. We modeled this error
process using simple Gaussian blurring. As in missing section augmentation, we
picked random $z$-locations up to five slices and applied a 2D Gaussian blur
filter either partially or fully. The standard deviation of Gaussian filter was
randomly sampled from the uniform distribution between zero and five pixels.

%-------------------------------------------------------------------------------
% Section 6. Mean affinity agglomeration
%-------------------------------------------------------------------------------
\section{Mean affinity agglomeration}
\label{mean_affinity}

Oversegmentation into supervoxels followed by agglomeration has been proposed as
a strategy for segmenting EM images \cite{jain2011learning, nunez-iglesias2013,
bogovic2013learned, nunez2014graph}.  In this approach, each pair of adjacent
supervoxels receives an agglomeration score, and the pair with the highest score
is greedily merged at each step. Previous work has emphasized learning of the
scoring function, often using hand-designed features as input.  We have found
that scoring a pair of supervoxels with a single hand-designed feature, the mean
affinity of all edges between the supervoxels, often produces good agglomeration
accuracy.  The analog of mean affinity for a boundary map is already used as a
feature in the GALA agglomeration package
~\cite{nunez-iglesias2013,nunez2014graph}, and was previously used to segment
natural images \cite{arbelaez2011}.

The rationale is that mean affinity smooths out noise in the affinity map that
could lead to merge errors.  Anecdotally, we have found that it is surprisingly
difficult to substantially outperform mean affinity agglomeration by learning
from GALA-type features. This is perhaps because the quality of convolutional
network output has improved so much in the years since GALA was introduced.

%-------------------------------------------------------------------------------
% Section 7. Neural Circuit Reconstruction Experiments
%-------------------------------------------------------------------------------
\section{Experiments}
\label{experiments}

\subsection{Dataset}

The SNEMI3D challenge provides a single labeled image stack of size
$1024\times1024\times100$ for training and the same-sized image stack for
testing. The voxel resolution is $6\times6\times29~nm^3$, which roughly amounts
to an anisotropy factor of 5. We further divided the training stack into top 80
slices for training and bottom 20 slices for validation.

\subsection{Model comparison}

We systematically examined the effect of our proposed data augmentation and
long-range affinity by comparing networks trained with different setups.
\texttt{aug0} refers to the nets trained with none of our proposed augmentation,
and \texttt{aug3} refers to those trained with all of them. Note that both
setups still include the basic types of data augmentation described in
Section~\ref{augmentation}: rotation, flip, warping, brightness and contrast
augmentations. Postfix \texttt{-long} is used to indicate whether the net was
trained with long-range affinity. By combining these setups, we
trained a total of four nets on the SNEMI3D training set, namely, \texttt{aug0},
\texttt{aug0-long}, \texttt{aug3}, and \texttt{aug3-long}.

Model selection and hyperparameter search were strictly performed on the
validation set. We have only submitted the result of \texttt{aug3-long} to the
SNEMI3D challenge leaderboard (Table~\ref{tab:SNEMI3D}). We subsequently
performed extensive quantitative comparison on AC3, a labeled image stack of
size $1024\times1024\times256$ that was made publicly available along with the
publication of Kasthuri et al.~\cite{kasthuri2015}. It should be noted that
although AC3 is a superset of the SNEMI3D test set, we used it only for the
post-challenge analyses after submitting our SNEMI3D results.

\subsection{Training procedures}

Our networks were trained using the binomial cross-entropy loss with
class-rebalancing. The network weights were initialized as described in He et
al.~\cite{he2015a}. We used the Adam optimizer~\cite{kingma2014}, starting with
$\alpha=0.01$, $\beta_1=0.9$, $\beta_2=0.999$, and $\epsilon=0.01$. The step
size $\alpha$ was halved when validation loss stopped decreasing, up to four
times. We used a single patch of size $160\times160\times18$ (i.e. minibatch of
size 1) to compute gradients at each training iteration. We trained our nets
until convergence using the Caffe deep learning framework~\cite{jia2014caffe}.
The total number of iterations for each training setup ranged from 500K to 700K.
Each training took about five days on a single NVIDIA Titan X Pascal GPU.

% We performed model selection and hyperparameter search on the validation set.
% For model selection, we sampled models at every 100K iteration and computed the
% variation of information (VI) metric on the validation set, selecting the best
% performing model for each training setup (\texttt{aug0}, \texttt{aug0-long},
% \texttt{aug3}, \texttt{aug3-long}). For the selected model, the segmentation
% threshold that produces an optimal (minimal) VI on the validation set was
% chosen.

\subsection{Postprocessing}

We used an edge-weighted graph implementation of watershed
algorithm~\cite{zlateski2015} to produce initial oversegmenation. We chose
parameters $T_\text{min} = 1\%$, $T_\text{max} = 80\%$, $T_\text{size} = (800,
20\%)$, and $T_\text{dust} = 600$. Note that relative percentiles computed from
the output affinity distribution were used instead of absolute parameter values.
We picked the best performing segmentation threshold optimized on the validation
set when generating our SNEMI3D submissions.

\subsection{Evaluation}

The SNEMI3D leaderboard measures segmentation quality based on the adapted Rand
F-score~\cite{rand1971,ignacio2015}. For the post-challenge analyses on AC3, we
adopted the variation of information (VI), an information theoretic metric, to
measure segmentation quality~\cite{nunez-iglesias2013,meila2007}. VI is
defined by
\begin{equation}
  VI(S,T) = H(S|T) + H(T|S),
\end{equation}
where $S,T$ are two segmentations to compare. Suppose that $S$ is a segmentation
produced by an automated method and $T$ is the ground truth. Then the
conditional entropy $H(S|T)$ measures oversegmentation errors (splitters), and
$H(T|S)$ measures undersegmentation errors (mergers).

% However, the Rand index-based
% metrics~\cite{rand1971} depends quadratically on the segment size, which is not
% desirable because the metric can be dominated by a few errors involving huge
% segments. We desired to minimize the overall error rate, so a metric that is
% less depdendent on the segment size is desirable.

%-------------------------------------------------------------------------------
% Section 8. Results
%-------------------------------------------------------------------------------
\section{Results}
\label{results}

% Main results.
\begin{table}[!b]
\caption{Results on the SNEMI3D challenge dataset.
\label{tab:SNEMI3D}}
\begin{center}
\begin{tabular}{|l|c|c|c|}
\hline
Group name & Rand error & Trainable parameters & Test-time augmentation \\
\hline
Ours (test-time aug.) & 0.02576 & 1.5M & 16 variants \\
Ours (test-time aug.) & 0.02590 & 1.5M & 8 variants \\
Ours (mean affinity aggl.) & 0.03332 & 1.5M & 1 variant \\
\textbf{** human values **} & \textbf{0.05998} & -- & -- \\
DIVE~\cite{zeng2017} & 0.06015 & 18M $\times$ 3 models & 16 variants $\times$ 3 models \\
IAL~\cite{beier2017} & 0.06561 & 35M & 20 variants \\
% IAL\_deprecated & 0.08040 & -- & -- \\
% Team Gala~\cite{nunez-iglesias2013} & 0.10041 & -- & -- \\
\hline
\end{tabular}
\end{center}
\end{table}

Table~\ref{tab:SNEMI3D} summarizes the SNEMI3D challenge leaderboard after our
submission of \texttt{aug3-long}. Our result was ranked first place on the
leaderboard, and strikingly, it has surpassed the human accuracy value provided
by the challenge organizer. To the best of our knowledge, this is the first
demonstration that a fully automated algorithm can surpass human accuracy in the
dense neural circuit reconstruction on any publicly available benchmark EM
dataset.

Comparing our method against the former leading
entries~\cite{zeng2017,beier2017} further demonstrates the effectiveness of our
approach. The number of trainable parameters of our ``deeply'' 3D net is an
order of magnitude smaller than the 2D~\cite{beier2017} or ``shallowly''
3D~\cite{zeng2017} convolutional nets~\footnote{Zeng et al.~\cite{zeng2017} used
an ensemble of three convolutional nets, each taking as input one, two, and
three consecutive slices. Their net has 3D convolution only in its initial
layers, where anistropy is maximal and thus the efficacy of 3D convolution would
be minimal.}. Our proposed mean affinity agglomeration is also quite remarkable
because it could achieve superhuman accuracy without the need for the costly
test-time augmentation (Table~\ref{tab:SNEMI3D}).

% Model comparison on AC3.
\begin{table}[!t]
\caption{Variation of Information (VI) measured on AC3.
\label{tab:AC3}}
\begin{center}
\begin{tabular}{|l|c|c|c|c|}
\hline
 & \texttt{aug0} & \texttt{aug0-long} & \texttt{aug3} & \texttt{aug3-long} \\
\hline
Baseline (1 variant) & 0.935 & 0.656 & 0.637 & \textbf{0.529} \\
Test-time augmentation (16 variants) & 0.607 & 0.578 & 0.552 & \textbf{0.500} \\
Mean affinity agglomeration (1 variant) & 0.568 & 0.554 & 0.546 & \textbf{0.513} \\
\hline
\end{tabular}
\end{center}
\end{table}

% % Test-time augmentation.
% \begin{figure}
% \begin{center}
% \includegraphics[width=1.0\textwidth]{test_time_aug.pdf}
% \end{center}
% \caption{The effect of test-time augmentation (16 variants) quantified on AC3.}
% \label{fig:test_time_aug}
% \end{figure}
%
% % Mean affinity agglomeration.
% \begin{figure}
% \begin{center}
% \includegraphics[width=1.0\textwidth]{mean_affinity.pdf}
% \end{center}
% \caption{The effect of mean affinity agglomeration quantified on AC3.}
% \label{fig:mean_affinity}
% \end{figure}

% AC3.
\begin{figure}
\begin{center}
\includegraphics[width=1.0\textwidth]{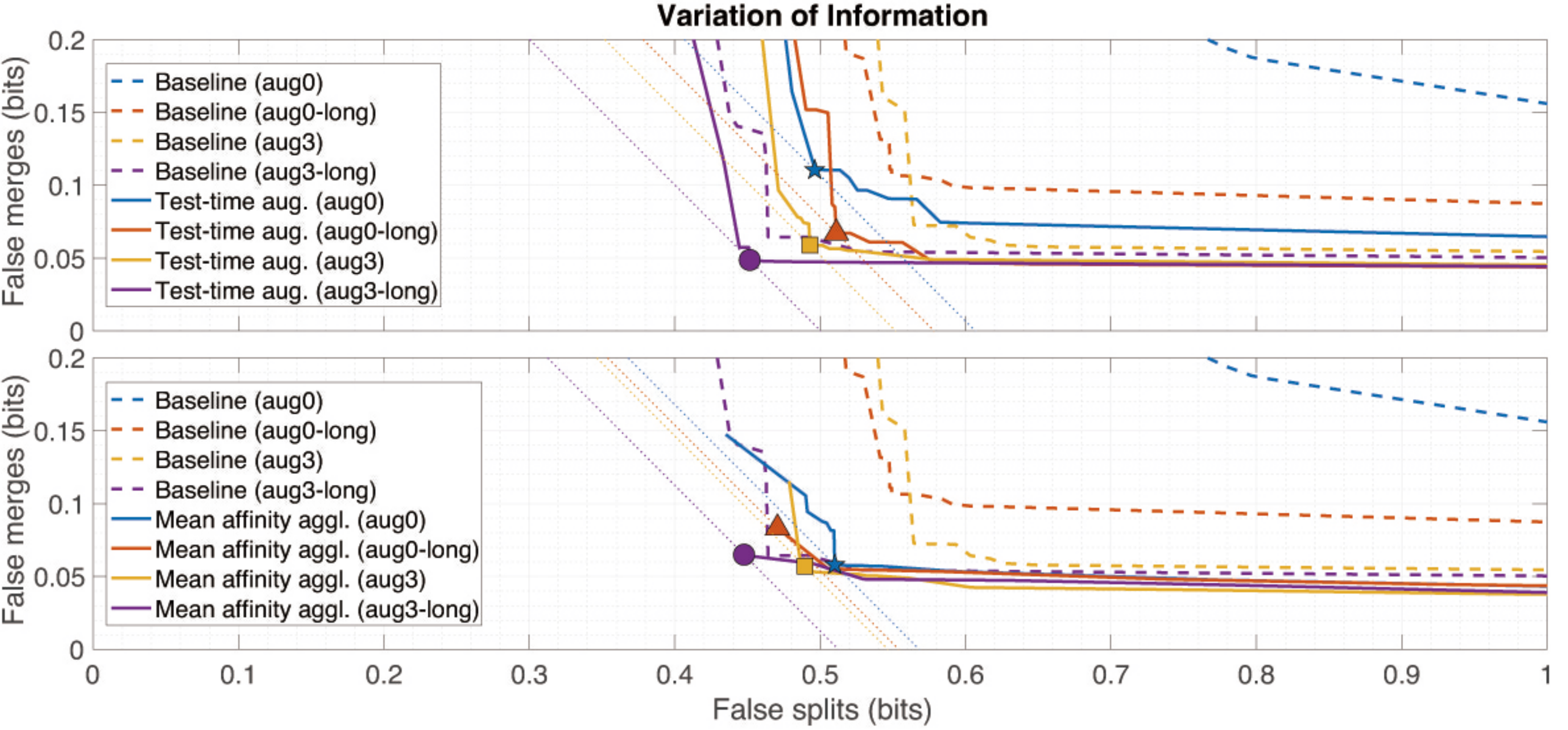}
\end{center}

\caption{The effect of test-time augmentation (top) and mean affinity
agglomeration (bottom) measured on AC3.}

\label{fig:AC3}
\end{figure}

\paragraph{Model comparison on AC3} More detailed quantitative comparison
between our models is shown in Table~\ref{tab:AC3} and Figure~\ref{fig:AC3}. On
the basis of this comparison, we can make the following claims. (1) Our proposed
data augmentation significantly improves model performance (\texttt{aug0} vs.
\texttt{aug3}, \texttt{aug0-long} vs. \texttt{aug3-long}). (2) Training with
long-range affinities substantially improves model performance (\texttt{aug0}
vs. \texttt{aug0-long}, \texttt{aug3} vs. \texttt{aug3-long}). (3) Test-time
augmentation boosts model performance at the expense of 8-16$\times$ inference
cost. (4) Mean affinity agglomeration is also very effective at boosting model
performance. Notably, mean affinity agglomeration was so effective that the
performance gap between the four models was more or less neutralized
(Table~\ref{tab:AC3} and Figure~\ref{fig:AC3}). Viewing from a different
standpoint, mean affinity agglomeration produces diminishing returns as the
underlying model keeps improving.

\begin{figure}
\begin{center}
\includegraphics[width=1.0\textwidth]{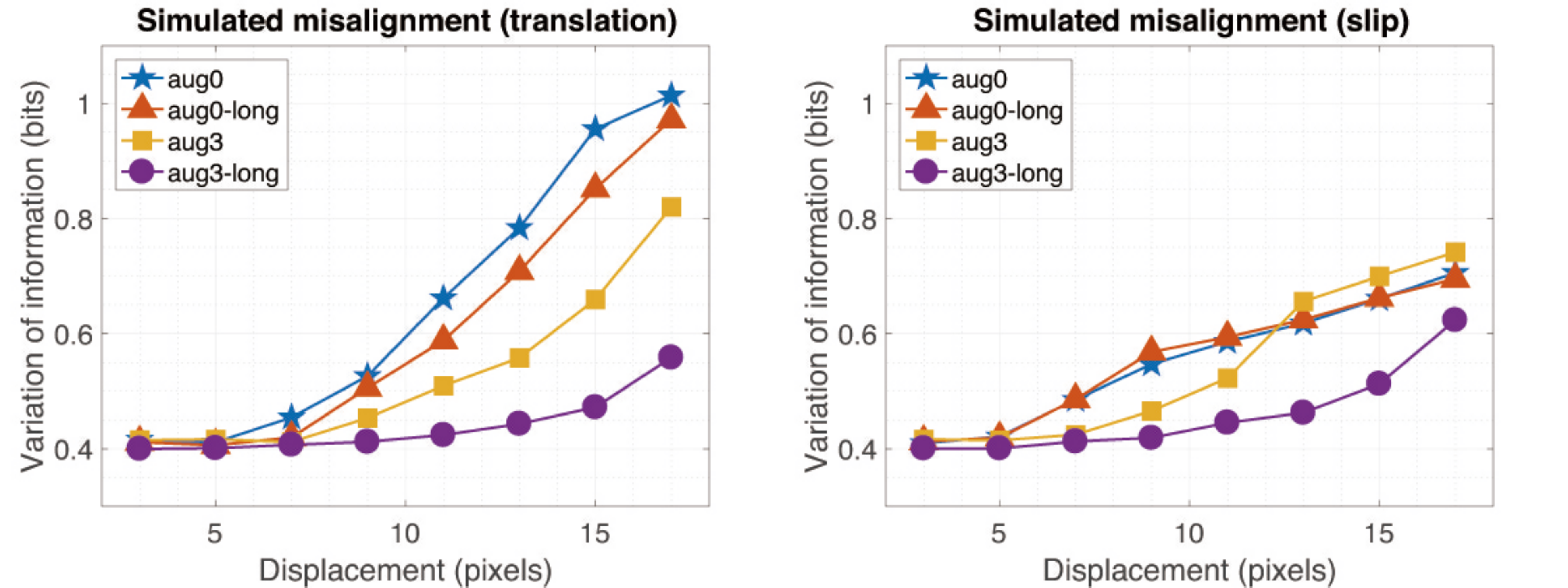}
\end{center}

\caption{Robustness to misalignment errors quantified on the validation set.
Left: \textit{translation}-type misalignment, right: \textit{slip}-type
misalignment. Here we used mean affinity agglomeration as a postprocessing.}

\label{fig:misalign}
\end{figure}

\paragraph{Effect of misalignment augmentation} We performed a couple of
additional analyses to demonstrate the effectiveness of our proposed data
augmentation. First, we systematically simulated the two types of misalignment
(\textit{translation} and \textit{slip}) on the validation set and examined how
robust the different models are. As we expected, the models trained with
misalignment augmentation were substantially more robust to the misalignment
errors (yellow and purple curves, Figure~\ref{fig:misalign}). One exception is
that \texttt{aug3} started to become worse than \texttt{aug0} and
\texttt{aug0-long} beyond a certain extent of \textit{slip}-type misalignment
(yellow curve in the right panel of Figure~\ref{fig:misalign}). We have not had
a chance to investigate why this particular exception occurred. Nevertheless, it
is apparent that the combined use of our proposed data augmentation and
long-range affinity makes the model superbly robust to the misalignment errors
(purple curves, Figure~\ref{fig:misalign}).

\begin{figure}
\begin{center}
\includegraphics[width=1.0\textwidth]{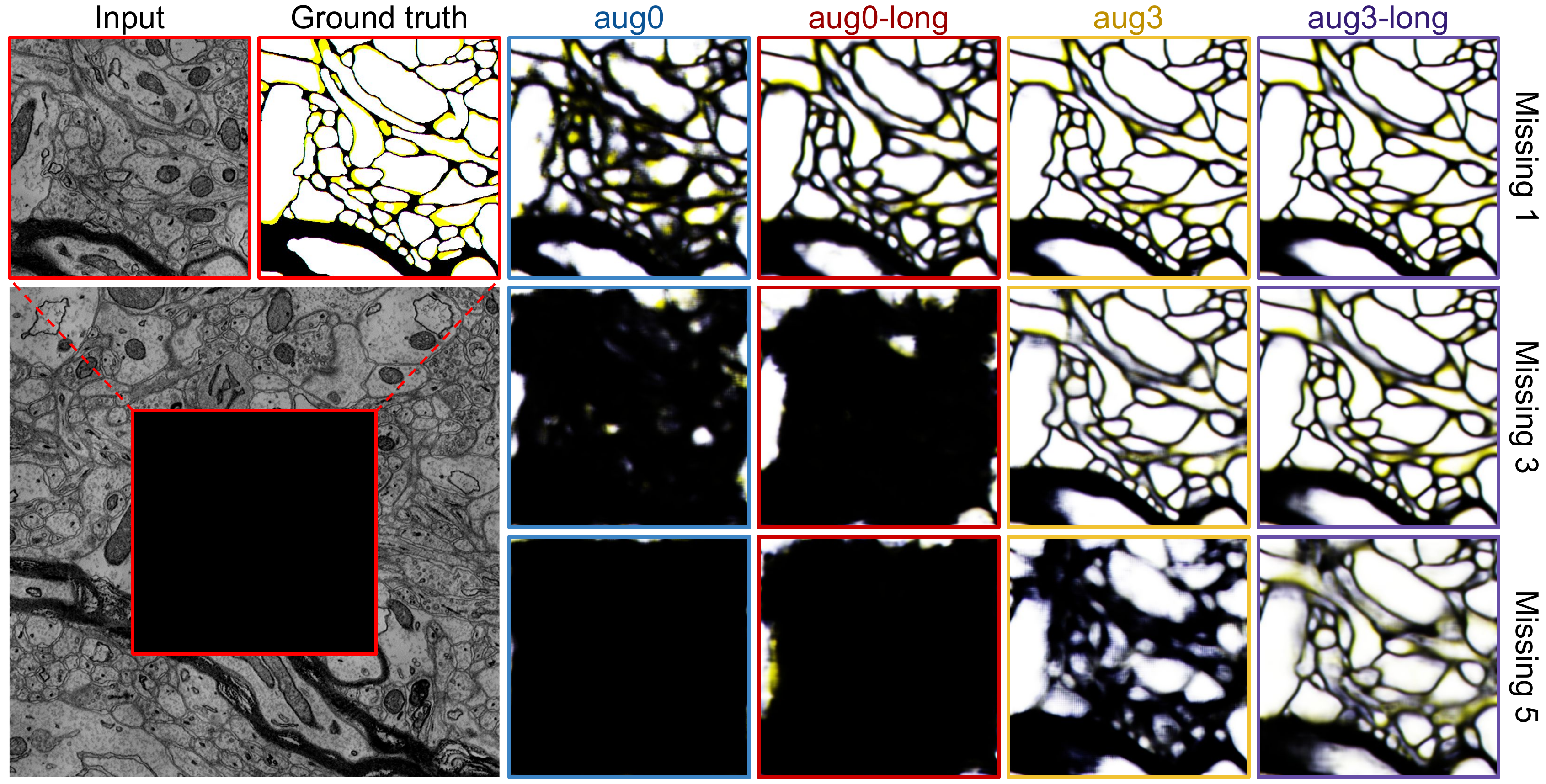}
\end{center}
\caption{Robustness to missing sections. Every result displays an affinity graph
as an RGB image (R: $y$-affinity, G: $x$-affinity, B: $z$-affinity).
Yellow-colored regions indicate the discrepancy between $z$-affinity and others.
Missing section augmentation enables nearly perfect completion of the missing
part when only a single section is missing (\texttt{aug3} and
\texttt{aug3-long}, top right corner).}
\label{fig:missing}
\end{figure}

\paragraph{Effect of missing section augmentation} To examine how robust the
different models are against missing section errors, we introduced one, three,
and five consecutive partial missing sections at the center of the validation
set. Figure~\ref{fig:missing} qualitatively illustrates each model's prediction
on the middle part of the missing sections. Interestingly, the models trained
without missing section augmentation (\texttt{aug0} and \texttt{aug0-long})
still managed to fill out the missing part to some extext when only a single
section was missing (top row, Figure~\ref{fig:missing}). However, both models
immediately failed at more than two consecutive missing sections. In contrast,
the models trained with missing section augmentation (\texttt{aug3} and
\texttt{aug3-long}) were substantially more robust even against multiple
consecutive missing sections. Again, the combination of our proposed data
augmentation and long-range affinity produced the best result
(\texttt{aug3-long}, the last column in Figure~\ref{fig:missing}).

% Misalignment.

% Missing section.

%-------------------------------------------------------------------------------
% Section 9. Discussion
%-------------------------------------------------------------------------------
\section{Discussion}
\label{discussion}

% % for discussion
% Note that postprocessing could still be helpful with nets for which accuracy has
% been sacrified for efficiency~\cite{meirovitch2016}.
%
% \paragraph{Large-scale sparse evaluation} Run length, etc.

\subsection{Failure modes}

What do the remaining errors look like? We observed that most of them fall into
one of the four categories: (1) errors caused by severe image defects, (2)
truely hard cases due to the limitations of serial section EM imaging (e.g.
extremely thin neurites that are parallel to the sectioning plane), (3) weakness
of mean affinity agglomeration on self-touching objects (e.g. dendritic spines
contacting the dendritic shaft from which they originated), and (4) object
classes that are largely underrepresented in the training set such as glial
cells surrounding blood vessels and soma-soma contacts.

We found that severe image defects are likely to cause bad misalignment errors,
which cannot be properly handled by the nearest neighbor affinity graph
representation. An obvious solution is to develop better image alignment
algorithms that are robust to such image defects. Another interesting
possibility suggested by our result (Figure~\ref{fig:missing}) is to completely
remove the image regions encompassing the defects and associated misalignment
errors, and then just let convolutional networks handle the missing sections.
Iterative refinement based on recursive/recurrent computation may be the key to
such a \textit{pattern completion} approach.

\subsection{Future directions}

A key ingredient that is still missing in the current automated pipeline for
neural circuit reconstruction is an automated way of detecting and correcting
the remaining errors. Meirovitch et al.~\cite{meirovitch2016} have recently
proposed a primitive rule-based error detection and a
flood-filling~\cite{januszewski2016} style approach to extend broken axons.
Supervised learning with deep neural networks may be applicable to the fully
automated error detection, which can potentially be useful for guiding focused
human proofreading.

Given automatically detected errors, a coupled automated error correction based
on another set of deep neural nets can also be conceivable. Flood-filling
network~\cite{januszewski2016} is a strong candidate for such tasks because it
can focus on a single erroneous object at a time and perform
perceptual/attentive computation that resembles the human way of correcting
errors. It is not even necessary that the error detector and corrector be
separate models.

%-------------------------------------------------------------------------------
% Acknowledgments
%-------------------------------------------------------------------------------
\subsubsection*{Acknowledgments}

We thank Barton Fiske of NVIDIA Corporation for providing us with early access
to Titan X Pascal GPU used in this research. We also thank Karan Kathpalia for
initial help with preliminary experiments on misalignment data augmentation and
Nicholas Turner for proofreading. Kisuk Lee was supported by a Samsung
Scholarship. This research was supported by the Intelligence Advanced Research
Projects Activity (IARPA) via Department of Interior/ Interior Business Center
(DoI/IBC) contract number D16PC0005. The U.S. Government is authorized to
reproduce and distribute reprints for Governmental purposes notwithstanding any
copyright annotation thereon. Disclaimer: The views and conclusions contained
herein are those of the authors and should not be interpreted as necessarily
representing the official policies or endorsements, either expressed or implied,
of IARPA, DoI/IBC, or the U.S. Government.

%-------------------------------------------------------------------------------
% References
%-------------------------------------------------------------------------------
\medskip
\small

\bibliography{main}
\bibliographystyle{unsrt}

\end{document}